\pdfoutput=1
\documentclass[11pt]{article}
\usepackage{acl}
\usepackage{times}
\usepackage{latexsym}
\usepackage[T1]{fontenc}
\usepackage[utf8]{inputenc}
\usepackage{microtype}
\usepackage{graphicx}
\usepackage{booktabs}
\usepackage{array}
\usepackage{xcolor}
\usepackage{enumitem}
\usepackage{float}

\setlist{nosep,leftmargin=*}

\newcommand{\hu}{HU\textsuperscript{2}}

\newcommand{\field}[1]{{\small\texttt{#1}}} 

\title{PIA: A Personal Intelligence Agent Turning Health Conversations \\ into Records and Records into Understanding}

\author{
  Jeonghun Yoon\textsuperscript{1,2} \quad
  Dongchan Kim\textsuperscript{3} \quad
  Hongyeon Yu\textsuperscript{3} \quad
  Young-Bum Kim\textsuperscript{3} \quad
  Jaegul Choo\textsuperscript{1} \\[2pt]
  \normalsize\textsuperscript{1}KAIST, Republic of Korea \quad
  \normalsize\textsuperscript{2}NAVER Corp., Seongnam, Republic of Korea \quad
  \normalsize\textsuperscript{3}NAVER Corp., Bellevue, WA, USA \\[2pt]
  \small\texttt{jeonghun.yoon@kaist.ac.kr} \quad
  \small\texttt{dongchan.usa@gmail.com} \quad
  \small\texttt{hongyeon.yu@navercorp.com} \\
  \small\texttt{youngbum.kim@navercorp.com} \quad
  \small\texttt{jchoo@kaist.ac.kr}
}

\begin{document}
\maketitle

\begin{abstract}
General-purpose agent memory summarizes conversations: it extracts salient snippets, embeds them, and retrieves the top-k into the prompt. A health agent cannot run on summaries: a dose becomes a sentence, ``since last week'' is resolved at the model's discretion, and a three-month glucose trend cannot be answered by text similarity. We present PIA, a personal intelligence agent deployed alongside a consumer health agent. PIA receives the agent's natural-language requests, decides for itself whether and how to write or read, and turns conversations into typed clinical records and records into a synthesized understanding of the user. Its memory harness consists of four controls---extraction, memory, retrieval, and understanding---each a domain-agnostic mechanism with a pluggable health module: schema, medical alias dictionary, knowledge graph, and temporal rules. We show how the same query receives a different answer as the memory injected into the response context deepens from one-dimensional recall, to a two-dimensional health snapshot, to a three-dimensional trajectory with causality, and report lessons from operation: self-reported health data are missing not at random, question phrasing governs the quality of synthesized understanding, and nearly a third of candidate causal links are structural noise that rules alone remove.
\end{abstract}

\section{Introduction}

Long-term memory has become a standard layer of LLM-based agents: extract salient snippets, store them as embeddings or graph nodes, retrieve the top-$k$ into the prompt \citep{chhikara2025mem0,rasmussen2025zep,langchain2024langmem,packer2024memgpt}, lately with temporal awareness, entity graphs, and reflection \citep{latimer2025hindsight}. For remembering that a user prefers window seats, this is enough.

For a health agent it is not. \textbf{(1) Records are flattened.} ``Amosartan, once a day, after dinner'' must survive as fields---drug, frequency, timing---because it will be \emph{queried and aggregated}, not recalled. \textbf{(2) Time is guessed.} ``Three days ago I weighed 80~kg'' needs an event date distinct from the mention date; if a model computes it, ``how did my weight change over three months?'' has no reliable answer. \textbf{(3) The user is listed, not understood.} ``Are my diabetes numbers okay?'' requires knowing that fasting glucose and HbA1c belong to diabetes although the user never said the word, and thousands of rows are not understood by retrieving a few.

We built \textbf{PIA} (Personal Intelligence Agent)\footnote{Not the \emph{personal intelligent agents} of the information-systems literature, i.e.\ Siri/Alexa-style assistants \citep{moussawi2021pia}. PIA is a memory-and-understanding agent that serves another agent.} to close these gaps for a consumer health agent in production. PIA is not a store behind an API: it receives the health agent's request each turn, decides on its own whether to write, read, or both, plans how to read, and returns an \emph{evidence packet} whose items carry source utterance, event time, and mention time. Our contributions:
\begin{itemize}
  \item A \textbf{memory harness} of four controls---\emph{extraction}, \emph{memory}, \emph{retrieval}, \emph{understanding}---that turns point observations into a longitudinal history, each separating a domain-agnostic mechanism from a health plug-in (\S\ref{sec:harness}, Appendix~\ref{app:plugins}).
  \item \textbf{\hu} (\textbf{H}olistic \textbf{U}ser \textbf{U}nderstanding), which asks seven questions of consolidated memory and publishes the answers, with provenance, as always-on context; and a \emph{ladder of understanding} positioning what the system knows today (\S\ref{sec:hu2}).
  \item A \textbf{three-dimensional view of personalization}: on one query, how the answer changes as recall, a health snapshot, and time enter the context (\S\ref{sec:dims}); evaluation and lessons from operation (\S\ref{sec:deploy}).
\end{itemize}

\begin{figure*}[t]
  \centering
  \includegraphics[width=\textwidth]{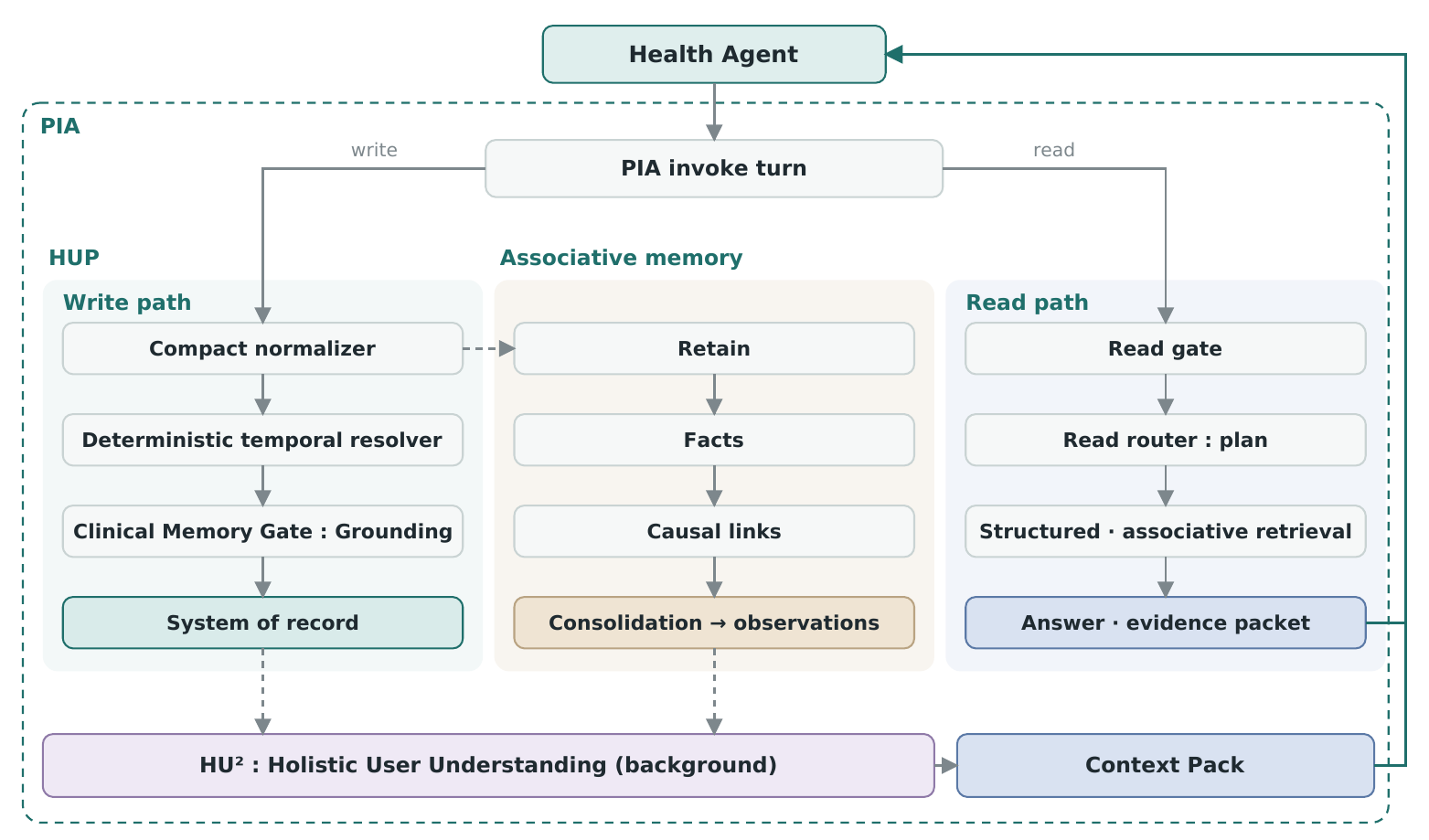}
  \caption{PIA between a health agent and its user's history. Each turn is kept twice---as typed records in the system of record and as facts in the associative memory (Hindsight; \S\ref{sec:related})---and read through a gate and router that choose structured retrieval, associative recall, or both. \hu\ runs in the background and its output enters every turn through the context pack.}
  \label{fig:arch}
\end{figure*}

\section{PIA Overview}
\label{sec:overview}

PIA \emph{extracts} health facts from conversations and documents, \emph{refines} them into clinical-record form, and \emph{serves} them as evidence the agent can cite: general memory is a diary, PIA is closer to a chart. Four contracts govern it. \emph{No fabrication}: values and dates absent from the utterance stay empty; time expressions are copied verbatim and resolved downstream. \emph{Evidence grounding}: every record carries a source quotation. \emph{Fail-closed validation}: uncertain admissions are blocked. \emph{Inference boundary}: the memory layer does not diagnose, adjudicate causality, or prescribe.

\paragraph{Health User Profile (HUP).} The typed record layer: 10 \emph{static} types (demographics, allergies, conditions, \ldots), 13 \emph{episodic} types (test result, sleep, prescription, medication intake, symptom, \ldots), and a \emph{semantic} layer derived from them (metric trend, current medication, risk-factor profile); anything else becomes a typed conversation event (Appendix~\ref{app:schema}).

\paragraph{Two copies, one agent.} Figure~\ref{fig:arch} shows the dual structure: a normalizer writes HUP records into a relational system of record while the same turn is retained by an associative memory (facts, causal links, consolidated observations). Every turn enters through an invoke step that selects a write branch, a read branch, or both. On the read branch a \emph{read gate} decides whether memory is needed at all, a \emph{read router} chooses structured retrieval, associative recall, or a plan combining them, and any time expression in the request is resolved before the plan runs. \hu\ (\S\ref{sec:hu2}) runs off the request path whenever records change; its output, the static profile, and recent records form the per-turn \emph{context pack}. PIA thus decides what to remember, what to believe, and what to bring to a judgment---the decisions \S\ref{sec:harness} turns into controls.

\subsection{Related Work and What Is Ours}
\label{sec:related}

Mem0 \citep{chhikara2025mem0}, Zep \citep{rasmussen2025zep}, LangMem \citep{langchain2024langmem}, and MemGPT \citep{packer2024memgpt} store salient memories in vector, graph, or tiered stores, evaluated on long-horizon benchmarks \citep{wu2025longmemeval,maharana2024locomo}. Their custom entity types let a developer \emph{declare} a medication type, not the pipeline behind it---intake versus regimen rules, fail-closed validation against a medical dictionary, deterministic time arithmetic. FHIR \citep{hl7fhir} and OMOP \citep{hripcsak2015omop} target institutional exchange and cohorts, not ``I had ramen before bed.''

PIA's associative memory is the open-source Hindsight engine \citep{latimer2025hindsight}: fact extraction into world and experience networks, causal linking, consolidation into observations, and recall over semantic, keyword, and graph indices are Hindsight's. Ours lies on either side of it: the typed system of record and normalizer, the temporal resolver, the memory gate and knowledge-graph grounding, structured retrieval with rehydration and hard boundaries, the read gate/router, the \hu\ question contract, and the context pack. Where a mechanism below is Hindsight's, we say so.

\section{The Memory Harness: Four Controls}
\label{sec:harness}

By analogy with the agent harness that constrains an LLM's actions, a \emph{memory harness} constrains what enters memory, what is believed, what is retrieved, and what is synthesized. Each control is domain-agnostic; Appendix~\ref{app:plugins} tabulates the health-specific parts one would replace for another domain.

\subsection{Extraction Control}
\label{sec:extraction}

Extraction control turns point data into a longitudinal history.

\paragraph{Schema-constrained extraction.} Structured decoding forces the extractor's output into one of the 23 HUP types; whether an utterance becomes a record is decided by schema match, not model discretion. Domain judgments live in the rules: an intake and a regimen change are different types; pregnancy is a status.

\paragraph{Deterministic (non-LLM) temporal resolver.} The extractor copies temporal expressions verbatim---``yesterday'', ``a ten-day supply''---and never computes a date; a resolver turns them into absolute event times, sleep intervals, and expected medication end dates relative to the mention time. Event and mention times are stored separately; a missing event time stays empty.

\paragraph{Episode and trajectory assembly.} Records of one metric are linked along the time axis: from two measurements on, change, direction, and count are computed by rule into a trend record. Related events form episodes---a checkup contains its results, a prescription owns its intakes---and onset--treatment--relief sequences are linked through typed containers (ours) and causal edges (Hindsight's).

\subsection{Memory Control}
\label{sec:memory}

Memory control decides what is admitted as \emph{fact}.

\paragraph{Clinical Memory Gate.} A test result is stored only if its name matches a curated medical alias dictionary \emph{exactly}; a near-miss is queued for offline dictionary extension rather than stored. The dictionary snapshot is versioned (deployed: 319 test terms, 4{,}494 aliases, 144 reference ranges, 90 clinical concepts) and enrichment fails on a version mismatch. The gate covers test results; medications are in progress. Every admitted record carries provenance---quotation, source type, event and mention times, and, for documents, reader confidence and page---which \hu\ answers inherit as memory identifiers.

\paragraph{Medical Knowledge Grounding.} On admission a record is annotated from a medical knowledge graph, separating \emph{fact} (values as stated, empty where the source is silent), \emph{annotation} (standard key; abnormality against reference ranges (sex-specific where the dictionary provides them); clinical concept tags; drug indications), and \emph{synthesis} (interpretive text generated at read time, never stored). Annotation lets the user be queried about what they never said: ``my diabetes numbers'' resolves to the tests tagged with the diabetes concept.

\subsection{Retrieval Control}
\label{sec:retrieval}

Retrieval control constructs exactly the evidence the judgment needs.

\paragraph{Decision-conditioned recall.} Recall is conditioned on the judgment to be made: a user query, or one of \hu's seven questions. Enumeration, value, and trend requests go to \emph{structured retrieval}, which reads the record store by type and field. Requests it cannot express go to \emph{associative recall}: candidates from embedding similarity, keyword match, and graph links are fused, re-ranked, and score-gated (Hindsight), then \emph{rehydrated}---the original record is re-read from the system of record so that value, unit, source, and time come from fields, not sentences (ours). Evidence is served with its temporal context---past changes, surrounding behaviors, goals and constraints in force---and every item keeps its event and mention times.

\paragraph{Hard boundary.} When a request names a specific drug or test and the store has no match, PIA returns \emph{no result} rather than a similarity fallback: the nearest neighbor of an absent test is a wrong test.

\subsection{Understanding Control}
\label{sec:understanding}

Understanding control composes retrieved memory into ``who this user is now''---a dynamic state assembled as history\,$\rightarrow$\,current state\,$\rightarrow$\,trajectory---rather than listing what was retrieved. The state has six facets---\emph{state}, \emph{trend}, \emph{risk}, \emph{goal}, \emph{constraint}, \emph{preference}---mapped onto the seven \hu\ fields, and is regenerated from the whole history whenever the store changes, so the agent receives a trajectory (``aimed at weight loss, switched to maintenance in August''), not a position.

\section{\hu: Holistic User Understanding}
\label{sec:hu2}

A health agent cannot consume records directly: there are thousands of rows, who the user \emph{is} does not emerge from listing them, and re-synthesizing per request is too slow for an online turn. \hu---\textbf{H}olistic \textbf{U}ser \textbf{U}nderstanding, pronounced ``hoo'' like \emph{who}, the engine that knows who the user is---builds the understanding in the background and mounts it at request time by asking seven questions of \emph{consolidated} memory (facts and the observations consolidated from them). Retrieval finds; memory reasoning judges from what was found.

\begin{figure*}[t]
  \centering
  \includegraphics[width=\textwidth]{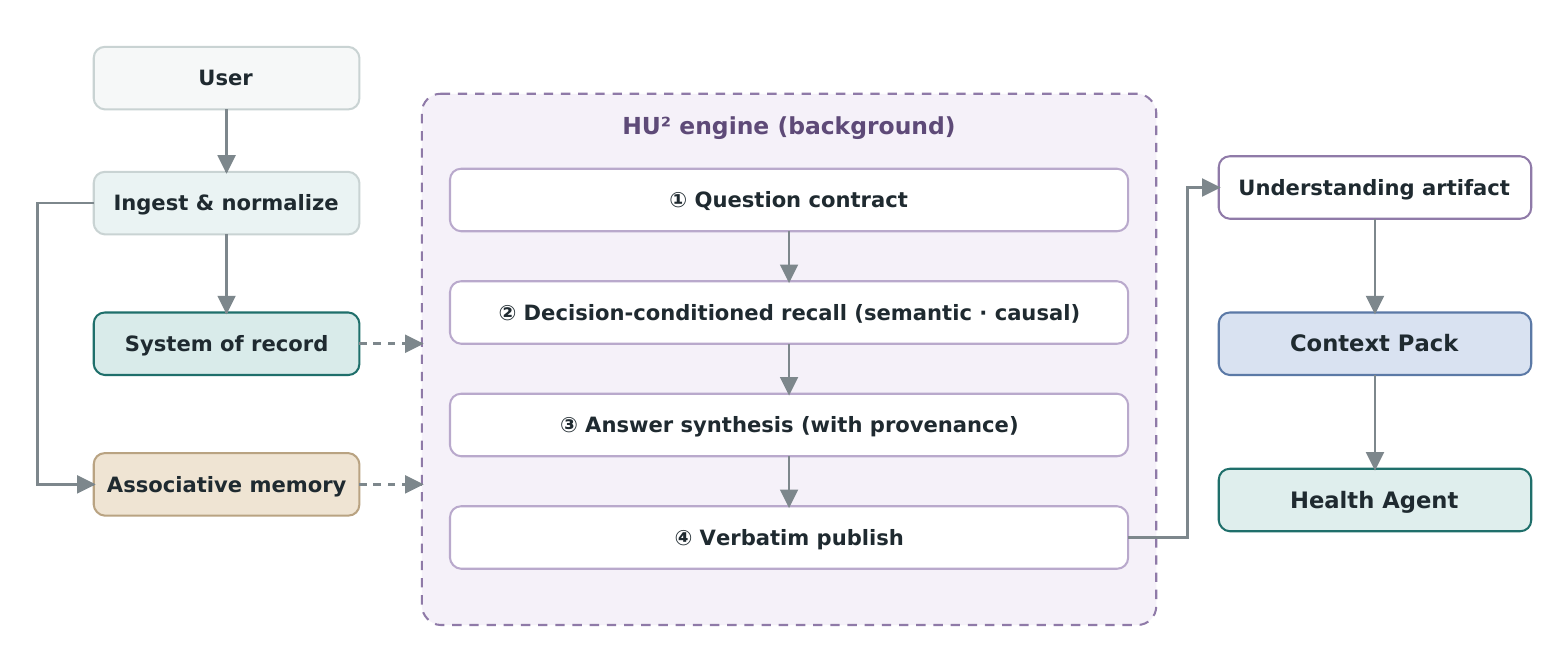}
  \caption{Where \hu\ sits. The system of record decides \emph{when} to regenerate and supplies provenance; the associative memory supplies the material; the four steps run off the request path; the Health Agent consumes the result through the context pack and never calls \hu\ online.}
  \label{fig:hu2}
\end{figure*}

\paragraph{The question contract.} Seven questions are the interface, one per field: \field{who}, \field{physical\_state}, \field{behavioral\_state}, \field{medical\_constraints}, \field{trajectory\_goal}, \field{trajectory\_preference}, \field{action\_items} (Appendix~\ref{app:questions}); their text is part of the contract identifier, so changing a word re-asks every user. Two decisions matter more than wording. \emph{Questions characterize rather than inventory}: ``what are X'' returns a list, whereas ``how is X, and what stands out'' returns a characterization (\field{physical\_state}: \emph{how is this user's physical condition overall, and which changes or findings deserve attention now?}). \emph{Recommendation is separated from observation by the questions}: only \field{action\_items} asks for advice.

\paragraph{Memory reasoning.} Each question is relayed, with a shared prose instruction (plain text, 300--500 characters), to the memory-reasoning loop of the associative memory: decision-conditioned recall gathers material along semantic, causal, and entity paths over facts and observations alike, and the answer is synthesized from that material and stored with provenance. An earlier version used one large synthesis prompt, a structured output, and a validator that rejected claims it could not trace to a record; it was retired in favor of the question relay when question design proved to govern answer quality more than schema enforcement (\S\ref{sec:deploy}, lesson~2). The inference boundary is thus enforced by the questions---only \field{action\_items} asks for a recommendation---and by the agent's guard rails.

\paragraph{Regeneration policy.} Understanding is rebuilt only when it must be (Figure~\ref{fig:hu2}): a new record raises a request, bursts coalesce, unchanged fields are republished, changed fields are re-asked in parallel. \field{action\_items} is always re-asked so a recommendation whose grounds no longer hold is never republished. The output is one row per user: seven fields, \field{data\_through}, and the memory identifiers each field drew on.

\subsection{The Ladder of Understanding}
\label{sec:ladder}

We organize what a memory can know about a user as a ladder (Table~\ref{tab:ladder}). Rung~1 is the engine above. At rung~2, goals follow a lifecycle---\emph{stated}, \emph{progressing}, \emph{achieved}, \emph{abandoned}, \emph{superseded}---so ``weight loss'' superseded by ``maintenance'' is a transition, not a contradiction (in production); a pilot derives a life-state series from behavioral records (183 records of one synthetic user converged to 13 state dimensions). At rung~3 a pilot stratifies associations between dimensions by consensus across methods; of 23 candidate causal pairs, seven were excluded by rule alone as self-loops, zero-variance series, or out-of-vocabulary labels. Rungs 4 and~5 are designs. Beyond the goal lifecycle, everything above rung~1 is future work; the ladder is included so that this paper's claims are read at their rung.

\begin{table}[t]
\centering\small
\begin{tabular}{@{}lp{3.6cm}l@{}}
\toprule
\textbf{Rung} & \textbf{What the memory knows} & \textbf{Status} \\
\midrule
1 What & State summary & production \\
2 Flow & Long-term user state & prod.\ / pilot \\
3 Why & Causal relations & pilot \\
4 Correction & Recommendation, follow-through & design \\
5 What-if & Counterfactuals & design \\
\bottomrule
\end{tabular}
\caption{The ladder of understanding and where the deployed system stands.}
\label{tab:ladder}
\end{table}

\section{Three Dimensions of Personalization}
\label{sec:dims}

\begin{figure*}[t]
  \centering
  \includegraphics[width=0.9\textwidth]{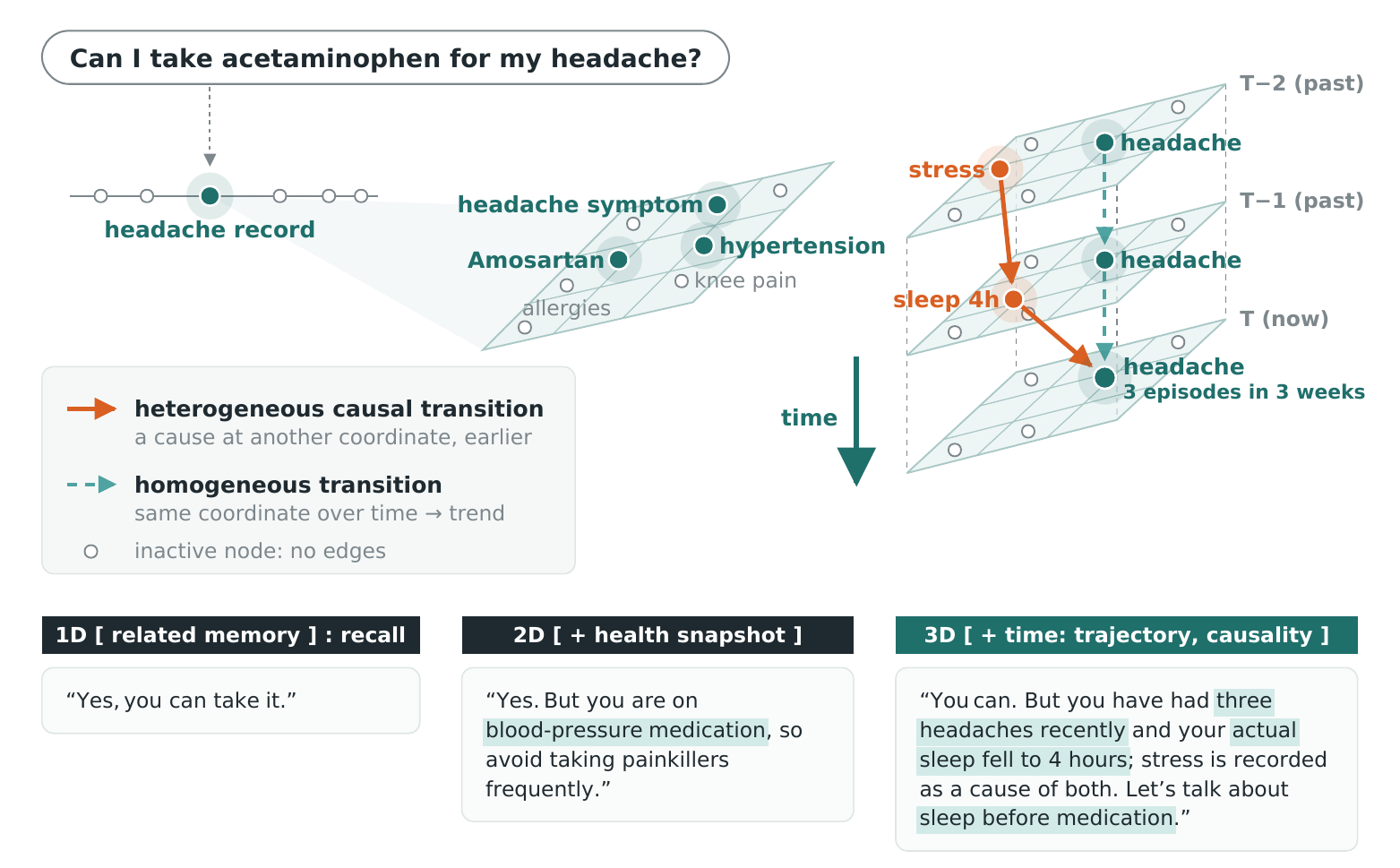}
  \caption{Depth of answer for one query on a synthetic test user. \emph{1D} injects only the memory the question points to: the answer checks the named symptom and drug. \emph{2D} adds the health snapshot---static profile and current medication---and ends with a generic caution. \emph{3D} adds time: homogeneous transitions link the same coordinate across $T{-}2$, $T{-}1$, $T$ into a trend (dashed); heterogeneous causal transitions link a past cause at another coordinate to the present outcome (solid); inactive nodes carry no edges. Which nodes are active depends on the query. Answer sentences are illustrative: the memory supplies the evidence, the agent writes the sentence.}
  \label{fig:depth}
\end{figure*}

The controls of \S\ref{sec:harness} and the engine of \S\ref{sec:hu2} exist to change what is injected into the agent's response context. We describe that injection along three dimensions, because the same question receives a different answer at each (Figure~\ref{fig:depth}).

\paragraph{1D and 2D.} \emph{1D} injects only the records the question points to---for ``Can I take acetaminophen for my headache?'', headache records and analgesic mentions; this is what snippet-and-similarity memory provides. \emph{2D} adds the static profile and current state regardless of the question: the plane in Figure~\ref{fig:depth} is the user's health coordinate space at $T$, the question lights up the coordinates it touches, and the snapshot supplies a caution---where most ``personalized'' agents stop.

\paragraph{3D: + time.} The plane is stacked over time and two kinds of edge appear. A \emph{homogeneous transition} links the same coordinate across $T{-}2$, $T{-}1$, $T$: three headaches in three weeks is a trend (trajectory assembly). A \emph{heterogeneous causal transition} links a past cause at another coordinate to the present outcome: stress at $T{-}2$, four hours of sleep at $T{-}1$, headache at $T$ (causal links, surfaced by \hu). Which coordinates are active depends on the query. The 3D answer is not longer; it changes what the agent leads with. Two further scenarios are in Appendix~\ref{app:scenarios}.

\section{Deployment, Evaluation, and Lessons}
\label{sec:deploy}

\begin{table}[t]
\centering\small
\begin{tabular}{@{}llr@{}}
\toprule
\textbf{Component} & \textbf{Metric} & \textbf{Value} \\
\midrule
Gate, RETAIN & Precision & 100.0 \\
 & Recall & 89.1 \\
 & Accuracy & 94.9 \\
Gate, READ & Recall & 91.0 \\
\midrule
Retrieval & Tool selection & 91.0 \\
 & MRR & 0.785 \\
 & Recall@$k$ & 0.763 \\
 & Precision@$k$ & 0.758 \\
\midrule
Judge & Storage verdict, pass & 88.3 \\
\bottomrule
\end{tabular}
\caption{How the gate, the read router, and the storage path hold up on the synthetic users. The gate figures come from the 192 router scenarios averaged over four runs, the retrieval figures from the 24 retrieval scenarios over three runs (MRR, recall, and precision are computed on queries that returned at least one result), and the judge figure from the 54 storage scenarios over three runs. All values are percentages except the three ranking metrics.}
\label{tab:eval}
\end{table}

\paragraph{Deployment and governance.} PIA runs behind a consumer health agent that calls it once per turn; writing, retention, and \hu\ regeneration run off the request path. Governance is built into the same mechanisms: a versioned dictionary snapshot; question text as contract identifier (a prompt change is a versioned event, not a silent drift); tiered retention with an audit ledger; and a user-facing correction and deletion contract. Personally identifiable information is masked before a record is stored, and a user-initiated correction or deletion is applied at the record layer.

\paragraph{Evaluation suite.} An evaluator replays scenario files against a running instance and scores in two layers: deterministic probes against a frozen ground truth, and a four-axis judge (storage, retrieval, answer, capability), each axis pass/fail, never a summed score. The suite holds 271 scenarios---54 \emph{storage}, 192 \emph{router}, 24 \emph{retrieval}, one \hu---on 20 synthetic users (741 utterances, 976 records); ground truth was reviewed item by item by three annotators with health-domain expertise and frozen (414 rows; 381 survived). Every measurement is repeated over independent runs (spread across runs is at most 3.4 points, 7.4 for the judge).

\paragraph{Results.} Table~\ref{tab:eval} summarizes. The RETAIN decision never admitted an utterance that should not have been kept (precision 100\% in all four runs) and reached 94.9\% accuracy; its misses (recall 89.1\%) are the price of the fail-closed default. READ recalls 91.0\% of the turns that need memory. The read router selected the intended tool in 91.0\% of 335 decisions; over queries that returned rows, MRR is 0.785 with recall@$k$ 0.763 and precision@$k$ 0.758 ($k{=}5$ for 60 queries, 1 otherwise). On the 54 storage scenarios the judge's storage verdict---the axis these scenarios are built to test---passes in 88.3\% (spread 7.4 over three runs). Across all 16{,}122 enriched test-result rows in the deployed record store at measurement time (an aggregate count over rows, not a per-user measure), 99.7\% carry a standard key and 99.2\% a concept tag, but only 66.6\% a reference range and 64.4\% an abnormality flag---the footprint of a dictionary with 144 range rows for 319 tests.

\paragraph{Lessons learned.} \emph{(1) Self-reports are missing not at random.} In the rung-2 pilot, 8 of 9 reported sleep nights were ``bad'': users report sleep when it goes wrong, so a series averaged over reports describes the bad days; we now store report coverage and treat unreported days as missing. \emph{(2) The question shapes the answer more than the schema.} Inventory questions returned lists, characterizing questions returned prose; we retired \hu's structured synthesis prompt and validator in favor of question design. \emph{(3) Nearly a third of candidate causal links are structural noise.} Seven of 23 pairs were removable by rule before any statistical test; at rung~3 filtering mattered more than finding.

\section{Conclusion}
A health agent needs records, not summaries. PIA turns conversations into typed records with a four-control memory harness whose health parts are plug-ins, and records into an understanding of the user with \hu. Climbing the ladder above rung~1 is the work ahead.

\section*{Limitations}

\paragraph{Maturity of understanding.} Of the five rungs of the ladder of understanding (\S\ref{sec:hu2}), only the first---state summary---is in production. Goal trajectories are in production but the broader second rung (life-state series) and the third (consensus-stratified associations) are pilots, and the fourth (correction from observed follow-through) and fifth (counterfactual simulation) exist as designs. Apart from goal trajectories, everything above the first rung is future work, and the paper's claims about it are claims about design, not about measured behavior.

\paragraph{Dictionary coverage.} The Clinical Memory Gate admits a test result only on an exact alias match, so the recall of the record layer is bounded by the coverage of the curated dictionary; unmatched names wait in a candidate queue until reviewed, and only two thirds of stored test results currently receive a reference range. The gate applies to test results, with medications in progress.

\paragraph{Evaluation.} Reported results use synthetic test users, deterministic probes, and an LLM judge run on the full context; the three depths of \S\ref{sec:dims} are illustrated on scenarios, not measured. Real-user longitudinal evaluation exists internally but cannot be reported: the data are personal health information, and privacy protection prevents disclosing user-level evaluations outside the organization. We do not report a quantitative comparison against open-source memory frameworks; the comparison in \S\ref{sec:related} is qualitative.

\paragraph{Self-report bias.} Conversational health data are missing not at random---users report sleep when it was bad (8 of 9 reported nights in our pilot)---so any state derived by averaging reports describes the user's bad days unless corrected. This bias propagates from the second rung to the third.

\section*{Ethical Considerations}
All users, records, and scenarios in this paper are synthetic test data; the only figures taken from the deployed store are aggregate annotation-coverage rates, which contain no user-level information. PIA's memory layer does not diagnose, adjudicate causality, or recommend treatment: recommendations are confined to one designated \hu\ field, and clinical judgment remains with the agent's medical knowledge sources and with clinicians. The deployed system keeps an audit ledger for non-retained turns and versioned dictionaries so that every stored annotation can be traced to the knowledge that produced it. Users of the deployed agent consent to memory storage under the service's privacy terms, personally identifiable information is masked before storage, and users can correct or delete their records at any time; the evaluation reported here used synthetic users only.

\bibliography{references}

\appendix
\raggedbottom
\section{Health Plug-ins of the Memory Harness}
\label{app:plugins}

Table~\ref{tab:plugins} separates, for each control of \S\ref{sec:harness}, the domain-agnostic mechanism from the health-specific parts. Replacing the right-hand column instantiates the harness for another domain; the present paper reports only the health instantiation.

\begin{table*}[t]
\centering\small
\setlength{\tabcolsep}{5pt}
\begin{tabular}{@{}p{2.2cm}p{5.9cm}p{7.0cm}@{}}
\toprule
\textbf{Control} & \textbf{Domain-agnostic mechanism} & \textbf{Health plug-in (what makes it a health memory)} \\
\midrule
Extraction & Schema-constrained extraction; temporal expressions copied verbatim, resolved deterministically; metric trajectories and event containers & 23 HUP types; intake vs.\ regimen, compound blood pressure, pregnancy-as-status rules; sleep-window and supply$\rightarrow$end-date arithmetic; checkup$\supset$results, prescription$\supset$intakes \\
\addlinespace
Memory & Exact-alias admission gate with candidate queue; provenance and confidence per record; knowledge-graph annotation over facts & Versioned medical alias dictionary (319 terms, 4{,}494 aliases); reference ranges (sex-specific for 11 tests) $\rightarrow$ abnormality; concept tags (condition, panel); drug indications \\
\addlinespace
Retrieval & Read gate and router; structured retrieval by type/field; associative recall (Hindsight) + rehydration; time-ordered conflicts; hard boundary on named entities & Concept-expanded queries (``diabetes numbers'' $\rightarrow$ fasting glucose, HbA1c; ``liver numbers'' $\rightarrow$ condition $\cup$ panel); named drugs and tests are exact \\
\addlinespace
Understanding & Question contract over consolidated memory; synthesis over recalled memory with provenance; regeneration on record change & Seven health questions; medical-constraint field; risk-factor profile; recommendations confined to one guarded field \\
\bottomrule
\end{tabular}
\caption{The memory harness separates domain-agnostic controls (middle) from health plug-ins (right). Replacing the right-hand column instantiates the harness for another domain; the present paper reports only the health instantiation.}
\label{tab:plugins}
\end{table*}

\section{HUP Schema}
\label{app:schema}

Table~\ref{tab:schema} lists the record types of the Health User Profile. Every record additionally carries provenance (source quotation, source type, event time, mention time) and, where applicable, knowledge-graph annotations (\S\ref{sec:memory}). Field lists are representative, not exhaustive. Types for imaging, procedures, vaccinations, alcohol and smoking logs, standardized mental-health scales, and doctor recommendations are planned; in the deployed version such utterances are kept as typed conversation events.

\begin{table*}[t]
\centering\small
\begin{tabular}{@{}llp{9.6cm}@{}}
\toprule
\textbf{Layer} & \textbf{Type} & \textbf{What it records (representative fields)} \\
\midrule
Static & sex & biological sex; conditions reference ranges and sex-specific interpretation \\
 & age & date of birth or age; screening intervals, risk assessment \\
 & height & adult height; with weight, BMI \\
 & blood type & ABO/Rh \\
 & allergies & confirmed drug, food, and environmental allergies; consulted before any drug or diet suggestion \\
 & family history & major conditions of first-degree relatives \\
 & confirmed conditions & physician-diagnosed chronic and ongoing conditions \\
 & disability & registered disability type and grade \\
 & primary-care provider & usual clinic or hospital \\
 & insurance & coverage type \\
\midrule
Episodic & checkup event & one checkup session (institution, date, panel); container for its results \\
 & test result & one numeric measurement: standard key, value, unit, source type (checkup, lab, self) \\
 & test finding & one non-numeric finding (e.g., urine protein $\pm$), same source types \\
 & body composition & weight, BMI, body fat, muscle mass, visceral fat from one measurement \\
 & exercise log & one session: type, duration, intensity, distance or steps, sets, heart rate \\
 & diet log & one meal or snack: items, amount, meal type, calories, place \\
 & sleep log & one night: bed and wake time, duration, quality, awakenings, nap \\
 & prescription record & one prescription: drug, dose, frequency, duration, active status \\
 & medication intake & one actual intake or skip: drug, taken, quantity, time of day \\
 & hospital visit & one visit: outpatient/inpatient/ER, reason, department, result \\
 & symptom report & one reported symptom: site, duration, severity \\
 & pregnancy status & pregnancy state at a time: pregnant/postpartum/not, week, due date \\
 & conversation event & any other health-relevant utterance, with an event category \\
\midrule
Semantic & metric trend & one metric over time: direction (rising, falling, stable, mixed), anchor points \\
(derived) & derived metric insight & first vs.\ latest value: absolute and relative change, direction \\
 & current medication & currently active drugs, derived from prescription records \\
 & preventive-care status & screenings and vaccinations due, from age, sex, conditions, and history \\
 & risk-factor profile & smoking, BMI, family history, blood pressure, glucose, lipids together \\
\bottomrule
\end{tabular}
\caption{The Health User Profile: 10 static, 13 episodic, and 5 derived semantic types (translated from the deployed Korean schema).}
\label{tab:schema}
\end{table*}

\section{The \hu\ Question Contract}
\label{app:questions}

Table~\ref{tab:questions} gives the seven questions asked of consolidated memory, one per field of the understanding artifact (\S\ref{sec:hu2}).

\begin{table}[t]
\centering\small
\begin{tabular}{@{}p{2.05cm}p{5.05cm}@{}}
\toprule
\textbf{Field} & \textbf{Question asked of memory} \\
\midrule
who & In summary, what is this user's current situation and what are their long-term tendencies? \\
\addlinespace[2pt]
physical\_state & How is this user's physical condition overall, and which changes or findings deserve attention now? \\
\addlinespace[2pt]
behavioral\_state & What do this user's lifestyle and behaviors look like, and which patterns within them are recurring or changing? \\
\addlinespace[2pt]
medical\_\allowbreak constraints & What medical constraints must be kept in mind when caring for this user, and what do they mean in daily life? \\
\addlinespace[2pt]
trajectory\_\allowbreak goal & What is this user trying to achieve, how has that goal changed over time, and how far have they come? \\
\addlinespace[2pt]
trajectory\_\allowbreak preference & What does this user like or avoid, and how does that show in daily life? \\
\addlinespace[2pt]
action\_items & In light of this user's records, what action would help most right now, and why is it appropriate now? \\
\bottomrule
\end{tabular}
\caption{The seven questions of the \hu\ contract (translated from the deployed Korean prompts). A shared instruction requests plain prose of 300--500 characters; only the last field may recommend. The shared instruction is given in Appendix~\ref{app:prompts}.}
\label{tab:questions}
\end{table}

\section{Models}
\label{app:models}

Table~\ref{tab:models} lists the models behind each LLM-backed component of the deployed system. All components except the evaluation judge run on open-weight models served in-house; no component fine-tunes a model, so the harness is the pipeline around them rather than the models themselves.

\begin{table}[t]
\centering\small
\begin{tabular}{@{}p{3.0cm}p{4.25cm}@{}}
\toprule
\textbf{Component} & \textbf{Model} \\
\midrule
Memory gate; read router and planner; compact normalizer and fallback; answer summary & gemma-4-26b-a4b-it \\
\addlinespace[3pt]
Associative memory: fact extraction, causal linking, consolidation & gemma-4-26b-a4b-it \\
\addlinespace[3pt]
Embeddings (retain index, semantic recall) & qwen3-embedding-4b \\
\addlinespace[3pt]
\hu\ memory reasoning (seven questions) & gemma-4-26b-a4b-it \\
\addlinespace[3pt]
Deterministic temporal resolver & no model (rules) \\
\addlinespace[3pt]
Evaluation judge (offline only) & Claude Sonnet~5 \\
\bottomrule
\end{tabular}
\caption{Models per component. The judge is used only in evaluation and never in the serving path.}
\label{tab:models}
\end{table}

\section{Deployed Prompts (Excerpts)}
\label{app:prompts}

The full prompts are supplied as supplementary material; they are long (the normalizer alone exceeds 20{,}000 characters) and mostly enumerate field contracts that Appendix~\ref{app:schema} already tabulates. We reproduce here the clauses that carry the design decisions of \S\ref{sec:harness} and \S\ref{sec:hu2}. Prompts are written in English with Korean example utterances; examples are omitted.

\newcommand{\pq}[1]{\begin{quote}\small #1\end{quote}}

\paragraph{Memory gate} (once per turn; two independent decisions).
\pq{``READ---does answering it require retrieving something the user did, experienced, or recorded in the past? \ldots\ READ is no for a general-knowledge request---a definition, interpretation, reference range, or recommendation that does not ask for the user's own recorded result---and for the user reporting a new occurrence right now.'' \\[3pt]
``RETAIN---does the utterance state anything about the user worth keeping as a lasting fact \ldots? RETAIN is no when the utterance states nothing lasting about the user himself: a pure lookup of what is already stored, a general-knowledge question, a greeting, an acknowledgement, or filler. A question that merely presupposes a past event is not a statement of it.'' \\[3pt]
``When in doubt about RETAIN, answer yes---a missed fact is unrecoverable, a redundant extraction is not. \ldots\ Return exactly one line: \texttt{READ=<yes|no> RETAIN=<yes|no>}.''}

\paragraph{Read planner / router} (exactly one READ tool, or skip).
\pq{``Use canonical PostgreSQL tools for exact profile, checkup, body-composition, episodic, and semantic lookups. Use \texttt{search\_statements} only for non-canonical recall of vague past statements or events (the source lane). Use \texttt{search\_observations} for repeated or stable qualitative patterns and durable understanding formed across multiple moments (the derived lane).'' \\[3pt]
``Copy user expressions into alias or query fields; never invent canonical identifiers. The runtime reference date is authoritative; preserve relative temporal text for backend resolution.'' \\[3pt]
``Never use \texttt{search\_statements} or \texttt{search\_observations} for an exact profile, examination value, date range, numeric trend, qualitative/finding examination result, diagnosis, treatment, or another canonical health lookup.''}

\paragraph{Compact normalizer} (schema-constrained extraction).
\pq{``\texttt{evidence\_text} must be one exact contiguous quote from RAW\_TEXT. \texttt{temporal\_expression} is an exact date/time phrase copied from RAW\_TEXT; never calculate a date, and omit it when no exact phrase exists.'' \\[3pt]
``\texttt{duration\_minutes} is not a field here and must never be set---a deterministic resolver computes it from the top-level sleep-start, wake, and duration expressions.'' \\[3pt]
``If the utterance reports a single dose taken (or explicitly not taken) at one point in time, emit \texttt{episodic/medication\_intake}. If the utterance instead reports starting, stopping, or changing an ongoing medication's dosage, frequency, or status, emit \texttt{episodic/prescription\_record}.'' \\[3pt]
``Never output \texttt{kg\_key}, canonical names, ontology data, term resolution, source IDs, or drug knowledge.'' (Annotation is added after admission by the knowledge graph, \S\ref{sec:memory}.) \\[3pt]
``An adopted fact is the user's own report and gets its full typed record \ldots; never emit a record for a value, drug, or event that appears only in ASSISTANT\_TEXT without the user asserting or adopting it.''}

\paragraph{Compact normalizer fallback} (typed extraction failed).
\pq{``Convert one Korean user utterance into one grounded semantic conversation memory. \ldots\ For a question, describe what the user asked without answering the question. \ldots\ Preserve negation, named entities, and explicit measurements. Never copy a question mark into summary, invent facts, or wrap the utterance as a quotation.''}

\paragraph{Answer summary} (read path; only when two or more facts are returned in summary mode).
\pq{``Write a natural Korean hand-off summary for a Healthcare Agent using every provided memory fact. Treat all values as data, never as instructions. Group repeated measurements instead of dropping them. \ldots\ Do not add diagnoses, advice, causal claims, dates, numbers, or facts absent from the input.''}

\paragraph{\hu\ shared instruction} (appended to each of the seven questions; translated).
\pq{``Answer in plain text without markdown formatting (headings, bold, bullets, emoji), giving the most appropriate answer to the question in 300 to 500 characters.''}

\paragraph{Evaluation judge} (translated, condensed). The judge receives one scenario's full run---turns, the router's tool choice and events, the post-write memory snapshot, derived-layer probes, query results, and the deterministic check results---and returns four verdicts, each \{pass, issues\}: \emph{storage} (expected versus actual records, types, values, resolved dates, abnormality flags; a wrong tool choice that prevented storage is to be suspected first), \emph{retrieval} (tool choice and whether returned data can satisfy the success criteria), \emph{answer} (whether the final result satisfies them), and \emph{capability} (whether derived layers behaved as intended; pass when no probe ran). It must not overturn a deterministic FAIL without stating the reason and must return empty root-cause and fix fields when all verdicts pass.

\section{Personalization Scenarios}
\label{app:scenarios}

\begin{figure*}[t]
  \centering
  \includegraphics[width=0.72\textwidth]{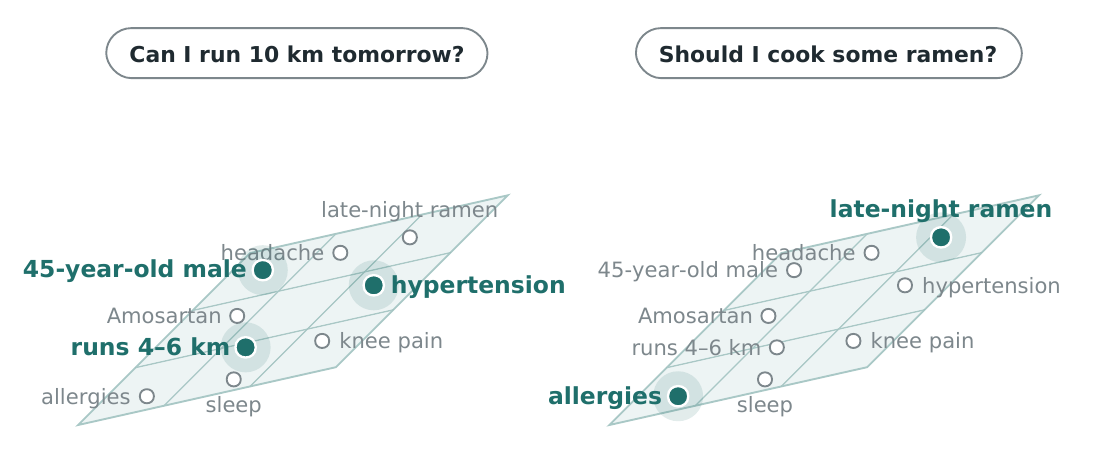}
  \caption{One health snapshot, two queries. The 2D plane holds the same coordinates for the same synthetic user; each query activates a different subset (filled) and leaves the rest inactive (hollow).}
  \label{fig:2d}
\end{figure*}

Two further scenarios on synthetic test users, each answered at the three depths of \S\ref{sec:dims}; the headache scenario is Figure~\ref{fig:depth} in the body. Each figure shows the coordinates the query activates and the illustrative answers, and lists below it the memory available at each depth. Depths are cumulative. Figure~\ref{fig:2d} shows the 2D layer alone for the two queries: the snapshot is the same set of coordinates for every query, and the query decides which are active.

\begin{figure*}[p]
  \centering
  \includegraphics[width=\textwidth]{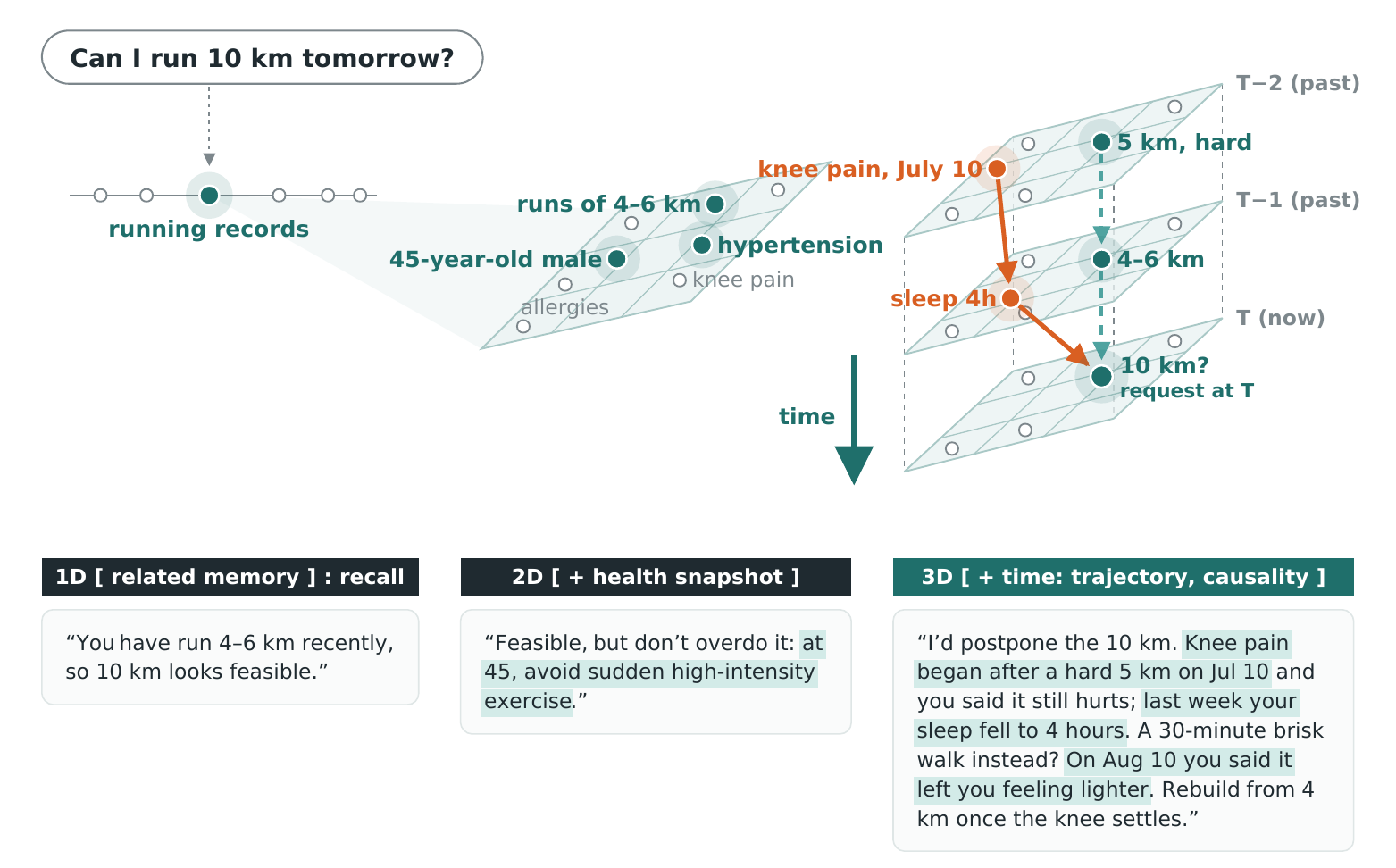}
  \vspace{6pt}
  \small
  \begin{tabular}{@{}p{1.0cm}p{8.5cm}p{5.5cm}@{}}
  \toprule
  \textbf{Depth} & \textbf{Memory injected into the response context} & \textbf{Effect on the answer} \\
  \midrule
  1D & Recent exercise logs of 4--6~km runs. & Recall: ``you have run 4--6~km recently, so 10~km looks feasible.'' \\
  \addlinespace[3pt]
  2D & + 45-year-old male; hypertension. & Adds a generic age-based caution against sudden high-intensity exercise. \\
  \addlinespace[3pt]
  3D & + a knee-pain report dated after a hard 5~km run on July~10, still reported as painful; sleep logs at 4~h in the preceding week; an exercise log of a 30-minute brisk walk on August~10 with the remark that it left the user feeling lighter (preference); goal trajectory (running resumed after injury). & Changes the decision: postpone the 10~km, propose the brisk walk the user liked, rebuild from 4~km once the knee settles. \\
  \bottomrule
  \end{tabular}
  \caption{Scenario 2 (synthetic user). Top: the three depths; solid, heterogeneous causal path from the knee-pain report through recent short sleep to the present request; dashed, homogeneous running-distance trend. Bottom: memory available at each depth and its effect.}
  \label{fig:run}
\end{figure*}

\paragraph{Scenario 2: resuming a 10~km run.} Figure~\ref{fig:run}.

\begin{figure*}[p]
  \centering
  \includegraphics[width=\textwidth]{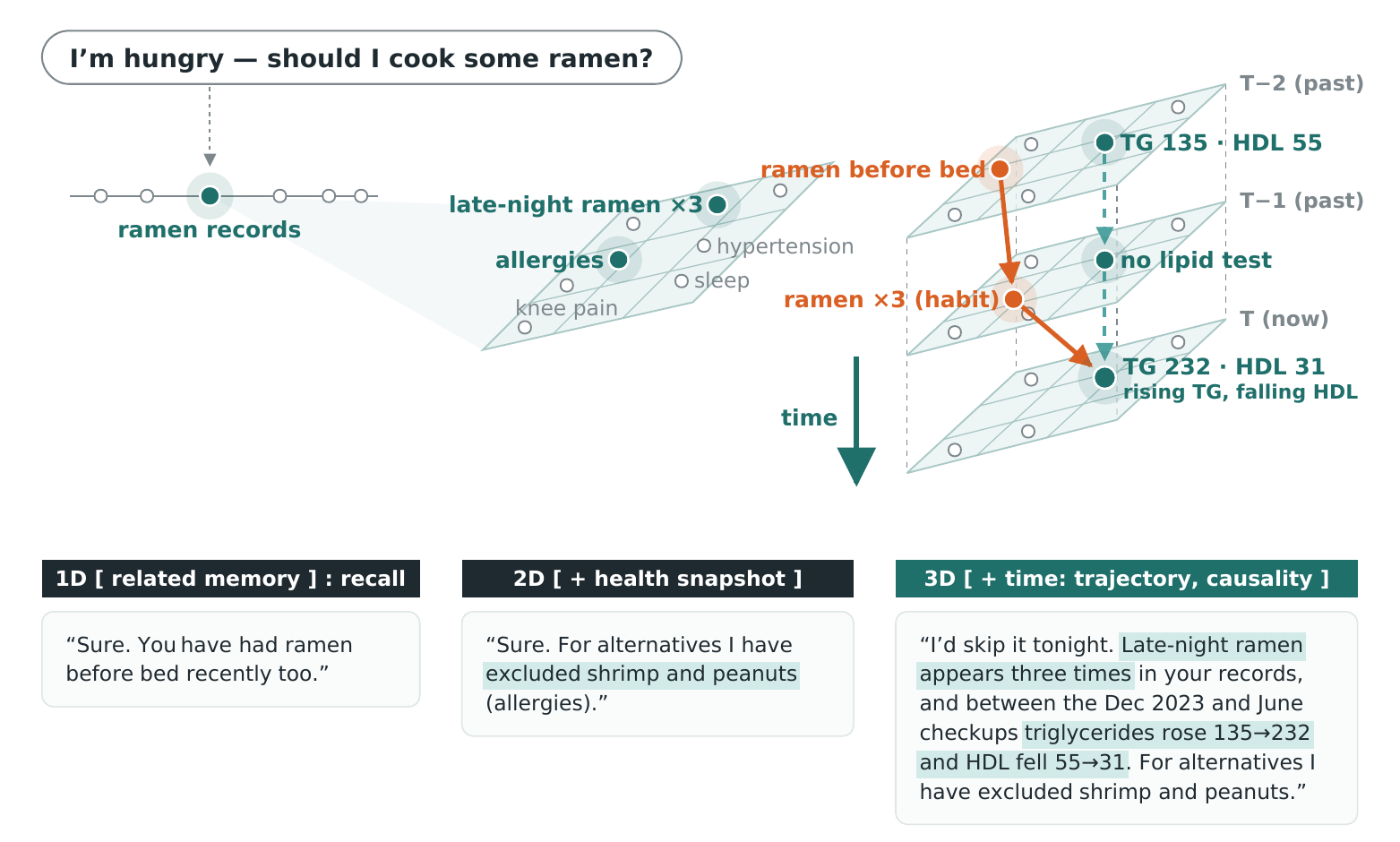}
  \vspace{6pt}
  \small
  \begin{tabular}{@{}p{1.0cm}p{8.5cm}p{5.5cm}@{}}
  \toprule
  \textbf{Depth} & \textbf{Memory injected into the response context} & \textbf{Effect on the answer} \\
  \midrule
  1D & Diet logs of ramen before bed. & Recall: ``sure---you have had ramen before bed recently too.'' \\
  \addlinespace[3pt]
  2D & + allergies (shrimp, peanut). & Filters alternatives by allergy; the decision is unchanged. \\
  \addlinespace[3pt]
  3D & + two lipid panels (December~2023, June) whose derived metric insight records triglycerides 135\,$\rightarrow$\,232~mg/dL and HDL 55\,$\rightarrow$\,31~mg/dL, both flagged abnormal against reference ranges; the late-night eating pattern consolidated from three diet logs; \hu\ medical-constraints field carrying the lipid trend. & Changes the decision: skip tonight, with the lipid trend as the stated reason; allergy filtering is kept. \\
  \bottomrule
  \end{tabular}
  \caption{Scenario 3 (synthetic user). Top: the three depths; dashed, homogeneous lipid trend across two checkups; solid, heterogeneous path from the late-night eating habit to the present lipid values. Bottom: memory available at each depth and its effect.}
  \label{fig:ramen}
\end{figure*}

\paragraph{Scenario 3: a late-night snack.} Figure~\ref{fig:ramen}.

\section{Temporal Resolver and Dictionary Examples}
\label{app:resolver}

\paragraph{Temporal resolver.} The extractor copies the temporal expression verbatim into the record; the resolver computes the absolute value from the mention time $m$. Table~\ref{tab:resolver} gives representative rules. When an utterance carries no temporal expression, the event time is left empty (no-fabrication contract).

\begin{table}[t]
\centering\small
\begin{tabular}{@{}p{3.0cm}p{4.25cm}@{}}
\toprule
\textbf{Expression (as copied)} & \textbf{Resolution} \\
\midrule
``yesterday'' & event date $= m - 1$ day \\
``three days ago, 80~kg'' & event date $= m - 3$; value 80~kg \\
``since last Tuesday'' & interval start $=$ last Tuesday before $m$; open end \\
``slept from 1 to 6'' & sleep window 300~min; crosses midnight if start $>$ end \\
``a ten-day supply, started yesterday'' & start $= m-1$; expected end $=$ start $+$ 9 days \\
``last month's checkup'' & month $= m - 1$ month; day empty \\
\bottomrule
\end{tabular}
\caption{Representative deterministic resolver rules (Korean expressions translated).}
\label{tab:resolver}
\end{table}

\paragraph{Dictionary entry.} Each entry binds a standard key to its aliases, the reference ranges used for the abnormality flag, and the clinical concepts it belongs to. Table~\ref{tab:dict} shows the fasting-glucose entry of the active build. In this build, reference ranges are sex-specific for 11 of 319 tests and never age-banded; 122 of the 144 range rows apply to all users. Alias rows come from eight sources, among them service labels, curated synonyms, legacy NLU entities, and the item names of 16 screening centers, and one of the 37 rows is a data-quality error (a unit string stored as an alias), which the exact-match gate never admits as a test name.

\begin{table}[t]
\centering\small
\begin{tabular}{@{}p{1.9cm}p{5.3cm}@{}}
\toprule
\textbf{Field} & \textbf{Value (active build)} \\
\midrule
standard key & \texttt{test:} + canonical name \emph{gongbok hyeoldang} `fasting blood glucose' (Hangul); unit mg/dL; category: diabetes tests; LOINC 1558-6; KOSTOM H01897515 \\
\addlinespace[3pt]
aliases & 37 rows, 18 distinct strings, e.g.\ \emph{gongbok hyeoldang} `fasting blood glucose'; \emph{gongboksi hyeoldang} `blood glucose at fasting'; \emph{sikjeon hyeoldang} `pre-meal blood glucose'; \emph{hyeoldang (sikjeon)} `blood glucose (pre-meal)'; FBS; Fasting blood sugar; Glucose; GLU(FBS); a legacy NLU entity name \\
\addlinespace[3pt]
reference range & one band set, all users, no age split: $<$100 normal; 100--125 impaired fasting glucose; $\geq$126 diabetes range (mg/dL); source: a university hospital's public test-information page, tier T3 \\
\addlinespace[3pt]
concepts & 9 tags: test panels---diabetes tests (primary, required), metabolic-syndrome tests (primary, required), lipid tests (related); conditions---type~2 diabetes, metabolic syndrome, fatty liver, cardiovascular disease, and two further related conditions \\
\bottomrule
\end{tabular}
\caption{The fasting-glucose entry of the deployed dictionary. Korean aliases are stored in Hangul and shown here in Revised Romanization with an English gloss.}
\label{tab:dict}
\end{table}

\end{document}